\documentclass{article}
\usepackage{ijcai24}

\usepackage{times}
\usepackage{soul}
\usepackage{enumerate}
\usepackage{makecell}
\usepackage[most]{tcolorbox}
\tcbuselibrary{breakable}
\usepackage{url}
\usepackage[hidelinks]{hyperref}
\usepackage[utf8]{inputenc}
\usepackage[small]{caption}
\usepackage{graphicx}
\usepackage{amsmath}
\usepackage{amsthm}
\usepackage{booktabs}
\usepackage{algorithm}
\usepackage{algorithmic}
\usepackage[switch]{lineno}

\usepackage{color}
\usepackage[inline,shortlabels]{enumitem}

\usepackage{amsmath,amsfonts,bm}

\def\eqref#1{equation~\ref{#1}}

\def\1{\bm{1}}

\def\vb{{\bm{b}}}

\def\ve{{\bm{e}}}
\def\vf{{\bm{f}}}

\def\vh{{\bm{h}}}

\def\vm{{\bm{m}}}

\def\vp{{\bm{p}}}
\def\vq{{\bm{q}}}

\def\vs{{\bm{s}}}
\def\vt{{\bm{t}}}

\def\vw{{\bm{w}}}

\def\mA{{\bm{A}}}

\def\mD{{\bm{D}}}

\def\mH{{\bm{H}}}

\def\mK{{\bm{K}}}

\def\mM{{\bm{M}}}

\def\mQ{{\bm{Q}}}
\def\mR{{\bm{R}}}

\def\mV{{\bm{V}}}
\def\mW{{\bm{W}}}
\def\mX{{\bm{X}}}

\DeclareMathAlphabet{\mathsfit}{\encodingdefault}{\sfdefault}{m}{sl}
\SetMathAlphabet{\mathsfit}{bold}{\encodingdefault}{\sfdefault}{bx}{n}

\def\gG{{\mathcal{G}}}

\def\gN{{\mathcal{N}}}

\def\gV{{\mathcal{V}}}

\newcommand{\R}{\mathbb{R}}

\usepackage{threeparttable}
\usepackage{multirow}
\usepackage{array}
\usepackage[normalem]{ulem}
\useunder{\uline}{\ul}{}
\usepackage{caption}
\usepackage{makecell}
\usepackage{amssymb} 
\usepackage{booktabs}
\usepackage{threeparttable}
\usepackage{amsthm}
\usepackage[figuresright]{rotating}

\title{HeMeNet: Heterogeneous Multichannel Equivariant Network for Protein Multi-task Learning}

\author{
    Rong Han$^1$ \and
    Wenbing Huang$^2$ \and
    Lingxiao Luo$^1$ \and
    Xinyan Han$^1$ \and
    Jiaming Shen$^1$ \and
    Zhiqiang Zhang$^3$ \and
    Jun Zhou$^3$ \and
    Ting Chen$^1$
}

\author{
Rong Han$^1$ \and
Wenbing Huang$^2$$^*$ \and
Lingxiao Luo$^1$ \and
Xinyan Han$^1$ \and
Jiaming Shen$^1$ \and
Zhiqiang Zhang$^3$ \and
Jun Zhou$^3$ \and
Ting Chen$^1$$^*$ \\
\affiliations
$^1$Tsinghua University, $^2$Renmin University of China, $^3$Ant Group \\
$^*$ Corresponding author
}
\newcommand{\datasetname}{Protein-MT}
\begin{document}

\maketitle



\begin{abstract}
Understanding and leveraging the 3D structures of proteins is central to a variety of biological and drug discovery tasks. While deep learning has been applied successfully for structure-based protein function prediction tasks, current methods usually employ distinct training for each task. However, each of the tasks is of small size, and such a single-task strategy hinders the models' performance and generalization ability. As some labeled 3D protein datasets are biologically related, combining multi-source datasets for larger-scale multi-task learning is one way to overcome this problem. In this paper, we propose a neural network model to address multiple tasks jointly upon the input of 3D protein structures. In particular, we first construct a standard structure-based multi-task benchmark called \datasetname{}, consisting of 6 biologically relevant tasks, including affinity prediction and property prediction, integrated from 4 public datasets. 
Then, we develop a novel graph neural network for multi-task learning, dubbed  \textbf{He}terogeneous  \textbf{M}ultichannel \textbf{E}quivariant \textbf{N}etwork (HeMeNet), which is E(3) equivariant and able to capture heterogeneous relationships between different atoms. Besides, HeMeNet can achieve task-specific learning via the task-aware readout mechanism. 
Extensive evaluations on our benchmark verify the effectiveness of multi-task learning, and our model generally surpasses state-of-the-art models.\footnote{Code Availability: \href{https://github.com/hanrthu/GMSL}{https://github.com/hanrthu/GMSL}}
\end{abstract}

\section{Introduction}
Proteins consist of one or more chains of amino acids, and they are vital in many biological systems. The 3D structure of a protein sets the foundation of its interaction with other molecules, which finally determines its functions. In recent years, learning-based methods have been applied widely to leverage the 3D structures of proteins for various tasks such as property prediction~\cite{wang2023deep}, affinity prediction~\cite{li2021structure}, rigid docking~\cite{ganea2022independent}, and antibody generation~\cite{kong2023endtoend}, owing to their superior efficiency and lower cost compared to those wet-lab experimental approaches.  A major part of learning-based methods resort to Graph Neural Networks (GNNs)~\cite{xu2018powerful}, which naturally encode the 3D structures of proteins by modeling atoms or residues as nodes and the connections in between as edges.
In addition, certain GNNs are geometry-aware and designed to capture the symmetry of E(3) transformations for better predictions~\cite{EGNN,huang2022equivariant}. 

\label{sec:dataconstruction}
\begin{figure}[t]
\begin{center}
\includegraphics[width=0.88\linewidth]{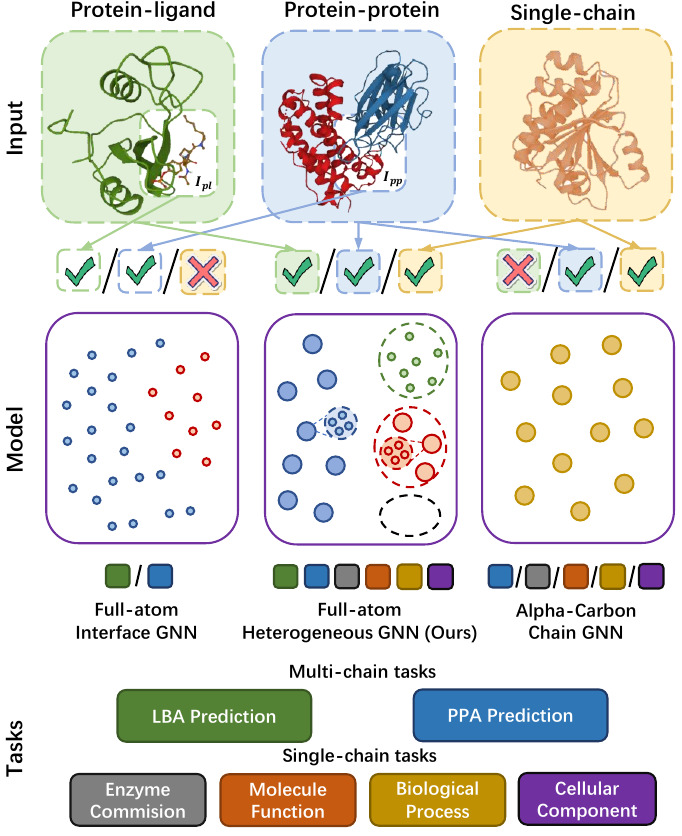}
\end{center}
\caption{\textbf{Comparison of different models with tasks}. Full-atom models (left) predict binding affinity with interface information; Alpha-Carbon models (right) predict protein functions with chain information. They need to be retrained for each task. HeMeNet (middle) supports various full-atom input information and predicts all six tasks simultaneously. We omit the edges for simplicity.}
\label{fig0}
\end{figure}

Despite significant progress in geometric-aware GNNs for protein tasks, existing methods usually employ one model for one task. A clear drawback of such a single-task training strategy is that the model should be re-trained for each new task. However,
structural datasets with annotations are often limited in size due to the expensive cost of acquiring protein 3D structures and labels via wet lab experiments, especially for affinity prediction. For example, in PDBbind~\cite{PDBbind2004}, only 2852 complexes for the Protein-Protein Affinity (PPA) are experimentally annotated.
Due to the sparsity of labeled structure samples, conducting model training on a single-task dataset of small size usually leads to defective performance and an inability to generalize. 

To deal with the sparsity of labeled data issues, some previous works leverage multi-task learning, which designs a model and trains it with multiple related tasks. Collecting samples with related tasks can bring more information to improve the performance~\cite{zhang2023biolip2,wang2022supervised}. 
However, most of these works are sequence-based~\cite{xu2022peer,capel2022proteinglue}, and the formulations are separated annotations for different samples. The work by\cite{capel2022multi} is a structure-based multi-task method, but it mainly focuses on residue-level interface prediction. 

Recent research shows that some protein properties potentially imply the protein's binding activity. For example, gene ontology would contain knowledge for protein-protein interaction~\cite{wang2022supervised}; Enzyme commission and gene ontology can provide molecular context for protein-ligand binding affinity (LBA)~\cite{zhang2023biolip2}. These indicate that some single-chain functions may benefit the prediction of complex-level affinity and vice versa. Motivated by this fact, we propose combining affinity and property prediction datasets in the framework of joint training. 

A key problem hindering structural data integration is the lack of appropriate models for various inputs and tasks. As shown in Figure \ref{fig0}, many affinity prediction models~\cite{li2021structure,kong2023generalist} utilize full-atom information at the binding interface to predict affinity, which loses the information of the whole chain. While many function prediction models~\cite{zhang2023protein,jing2021learning} utilize alpha-carbon to predict chain-level functions, which loses the detailed atom interaction information for affinity prediction.

In this paper, we propose a structure-based multi-task learning paradigm. We use a heterogeneous full-atom model to deal with multiple tasks upon various 3D protein inputs. 
Nevertheless, accomplishing structural multi-task training is challenging.
The first challenge is that there is no available benchmark. The ideal benchmark should cover a sufficient range of data and biologically related task types, with a fully labeled test set to compare how a model performs on the same input for different task outputs. 
The second challenge is that it is nontrivial to design a generalist model that is capable of processing the complicated 3D structures of input proteins of various types, including single-chain, protein-protein, and protein-ligand, and it should perform well across different tasks, including protein affinity and property predictions. 
By achieving the structure-based full-atom protein multi-task learning, we make the following contributions:
\begin{itemize}[leftmargin=12pt]
    \item To the best of our knowledge, we are the first to propose the concept of structure-based protein multi-task learning. We carefully integrate the structures and labels from 4 public datasets with our proposed standard process and construct a new benchmark named \textbf{Protein} \textbf{M}ultiple \textbf{T}asks (\textbf{Protein-MT}), which consists of 
    6 representative tasks upon 3 different types of inputs.
    \item We propose a novel model for protein structure learning, dubbed \textbf{He}terogeneous \textbf{M}ultichannel \textbf{E}quivariant Network (\textbf{HeMeNet}), which is E(3) equivariant and able to capture various relationships between different atoms owing to the heterogeneous multichannel graph construction of proteins. Additionally, we develop a task-aware readout mechanism by associating the output head of each task with a learnable task prompt for different tasks.
    \item For the experiments on \datasetname{}, HeMeNet surpasses other state-of-the-art methods in most tasks under both the single-task and multi-task settings. Particularly on the LBA and PPA tasks, we find that the multi-task HeMeNet is significantly better than its single-task counterpart.
\end{itemize}
\section{Related Works}

\paragraph{Protein Interaction and Property Prediction}
Predicting the binding affinity and properties for proteins with computational methods is of growing interest \cite{WangHuiwen2022,zhao2020literature}.
Previous research learns protein representations by information different forms, most of which take amino acid sequence~\cite{Alley2019,RaoRoshan2019}, multiple sequence alignment~\cite{rao2021msa} or 3D structure~\cite{IEConv,zhang2023protein} as input.
Many works encode the information of a protein’s 3D structure by GNNs~\cite{Vladimir2021,zhang2023protein}.
\cite{li2021structure} take full-atom geometry at the interaction interface, and~\cite{zhang2023protein} take residue-level geometry of the protein for property prediction.
Our method utilizes full-atom geometry on the whole protein to address affinity and property prediction tasks together.

\paragraph{Equivariant GNNs}
Many equivariant GNNs have emerged recently with the inductive bias of 3D symmetry, modeling various tasks including docking, molecular function prediction and sequence design~\cite{Nathaniel2018,Gasteiger2020,EGNN,SEGNN}. To empower the model with the ability to handle the complicated full-atom geometry, some models design multi-channel equivariant message passing for atom sets, such as GMN~\cite{huang2022equivariant} and dyMEAN~\cite{kong2023endtoend}.
We propose a powerful heterogeneous equivariant GNN capable of handling various incoming message types.
\begin{figure*}[t]
\begin{center}
\includegraphics[width=0.85\linewidth]{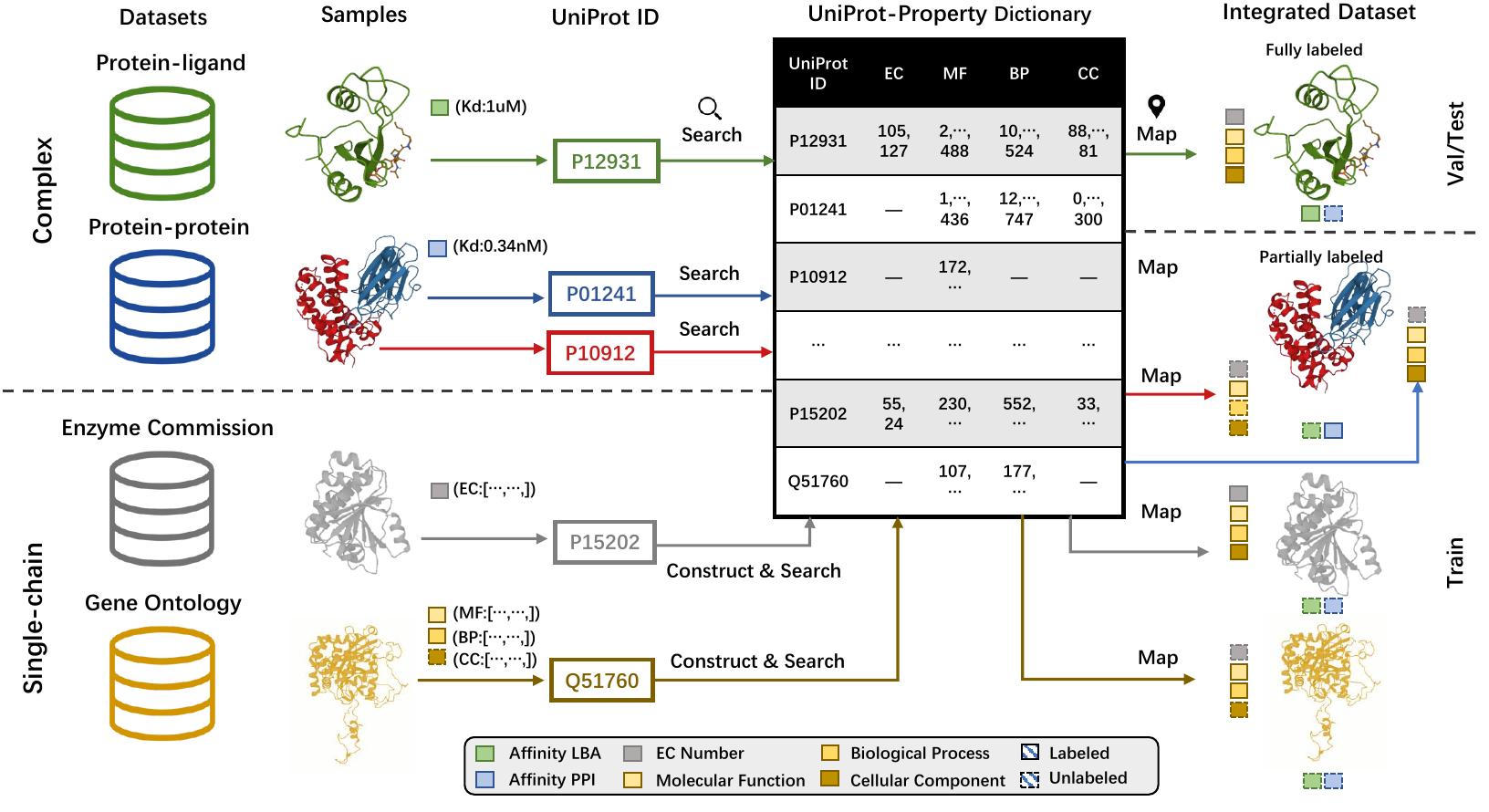}
\end{center}
\caption{\textbf{Construction of \datasetname{}}. We first extract the UniProt ID for each chain and construct a UniProt-Property dictionary to map the UniProt ID of each protein chain with EC and GO-MF, GO-BP, GO-CC labels annotated in the EC and GO datasets. With this dictionary, we can extract each chain's UniProt ID and map it with its labels. The complex with one affinity label and all property labels for each chain is defined as fully-labeled. We take most of the fully-labeled data for val/test and most of the partially labeled data for training.}
\label{fig1}
\end{figure*}

\paragraph{Protein Multi-Task Learning}
Multi-task learning takes advantage of knowledge transfer from multiple tasks, achieving a better generalization performance.
In the field of protein, several works leverage multi-task learning on the task of interaction prediction and property prediction, most of which are for sequence-based models.~\cite{shi2023enzyme} design three Enzyme Commission number related hierarchical tasks to train the model.
~\cite{wang2023multilevel} introduce a multi-task protein pre-training method with prompts.
~\cite{xu2022peer,capel2022proteinglue} are sequence-based multi-task benchmarks for protein function prediction. ~\cite{capel2022multi} improves the protein interaction interface prediction by structural multitask auxiliary learning.
To the best of our knowledge, we are the first to combine structure-based interaction prediction and property prediction in a multi-task setting.

\section{New Dataset: Protein-MT}
\label{sec:dataconstruction}


Based on the observation that protein property and binding affinity tasks may benefit each other, we construct a new dataset called Protein Multiple Tasks (\datasetname{}) for protein multi-task learning. 
\datasetname{} is composed of different types of tasks on 3D protein structures: the prediction of Ligand Binding Affinity (\textbf{LBA}) and Protein-Protein Affinity (\textbf{PPA}) based on two-instance complexes and the prediction of Enzyme Commission (\textbf{EC}) number and Gene Ontology (\textbf{GO}) terms based on single-chain structures. Particularly, the LBA and PPA tasks originated from the PDBbind database~\cite{PDBbind2004} aim at regressing the affinity value of a protein-ligand complex and protein-protein complex, respectively. The EC task is constructed by \cite{gligorijevic2021structure} to describe the catalysis of biochemical reactions consisting of samples, each with 538 binary-class labels. The GO task aims to predict the hierarchically related functional properties of gene products~\cite{gligorijevic2021structure}: Molecular Function (MF), Biological Process (BP), and Cellular Component (CC). We treat the prediction of MF, BP and CC as three individual tasks, resulting in six different prediction tasks in total.

One key difficulty in integrating these tasks from their sourced datasets is that samples from one task may lack the labels for other tasks. It is crucial to obtain samples with a complete set of labels across tasks for the training and evaluation of multi-task learning methods.
As shown in Figure \ref{fig1}, we propose a standard matching pipeline that enables us to transfer the labels between EC and GO, and assign EC and GO labels for the chains of complexes in LBA and PPA as well (it is impossible to conduct the inverse direction since it is meaningless to assign LBA or PPA for those single chains in EC and GO).
Specifically, we utilize the UniProt ID to uniquely identify a protein chain\footnote{The UniProt dataset is the world's leading non-redundant protein sequence and function dataset and it identifies proteins by their UniProt IDs.}.
We first obtain the UniProt IDs of all protein chains in \datasetname{} from Protein Data Bank \cite{berman2000protein}. For each UniProt ID, we then determine the EC and GO properties based on the labels of the corresponding chains in the EC and GO datasets, resulting in a UniProt-Property dictionary.
With this dictionary, for a chain missing EC or GO labels (e.g., a chain of a complex in LBA and PPA), we can supplement the missing labels by searching the UniProt-Property dictionary by its UniProt ID to retrieve any known EC and GO labels.
We define a complex (from either LBA or PPA) as fully-labeled if the complex has one affinity label (LBA or PPA) and four function labels for each of its chains. After our above matching process, we formulate the train/validation/test split in terms of the chain-level sequence identity through the alignment methods commonly used in single-chain property prediction tasks \cite{gligorijevic2021structure}. For more details of the construction process and dataset statistics, please refer to Appendix~\ref{appendix_a}.



\section{Methodology}
In this section, we first introduce our heterogeneous graph representation and the multi-task formulation in Section~\ref{sec:graph-representation}. Then, we design the architecture of the proposed HeMeNet in Section~\ref{sec:hemenet}, which consists of two key components: heterogeneous multi-channel equivariant message passing and task-aware readout.

\subsection{Heterogeneous Graph Representation and Task Formulation}
\label{sec:graph-representation}
The input of our model is of various types. It could be either a two-instance complex (protein-ligand for LBA and protein-protein for PPA) or a single chain (for EC and GO). Here, for consistency, we unify these two different kinds of input as a graph $\gG$ composed of two sets of nodes $\gV_{r}$ and $\gV_{l}$. For the LBA complex input, $\gV_{r}$ and $\gV_{l}$ denote the receptor and the ligand, respectively, while for the PPA complex and single-chain input, $\gV_{r}$ refers to the receptor protein chain and $\gV_{l}$ becomes the corresponding binding protein chain. And for function prediction tasks,  $\gV_{r}$ refers to the protein chain and $\gV_{l}$ becomes an empty set, as shown in the middle of Figure \ref{fig0}.  We associate each node $v_i$ with the representation $(\vh_i, \vec{\mX}_i)$, where $\vh_i\in\R^d$ denotes the node feature and it is initialized as a learnable residue embedding, $\vec{\mX}_i\in\R^{3\times c_i}$ indicates the 3D coordinates of all $c_i$ atoms within the node. As for edge construction, we include various types of edges. In detail, for residue nodes, we allow $R$ heterogeneous types of edge connections including sequential edges of different distances ($d = \{-2, -1, 1, 2\}$), self-loop edges, and spatial edges; for single-atom nodes from small molecules, only spatial edges are created.
We present a simplified example from the LBA task in Figure~\ref{fig2}, where we only draw a few nodes and omit the self-loop edges except for the central node for simplicity. Overall, we obtain a full-atom heterogeneous graph representation $\gG$ for each input.

\paragraph{Task Formulation}
Given a full-atom heterogeneous graph $\gG$, our goal is to design a model $\vp=f(\gG)$ with multiple-dimensional output $\vp$ that is able to predict the complex-level affinity and chain-level functional properties simultaneously. By making use of our proposed dataset \datasetname{}, we train the model with a partially labeled training set and test it on the fully-labeled test set.  Notably, the prediction should be invariant with regard to E(3) transformation (rotation/reflection/translation) of the input coordinates. To do so, we will formulate an equivariant encoder plus an invariant output layer in our model, detailed in the next subsection. 

\subsection{HeMeNet: Heterogeneous Multi-channel Equivariant Network}
\label{sec:hemenet}

To better cope with the 3D structures of different types for different tasks, we propose a heterogeneous multi-channel E(3) equivariant graph neural network with the ability to aggregate different relational messages. After several layers of the message passing, the node representations are transformed into task-specific representations by a task-aware readout module, generating appropriate complex-level and chain-level predictions via different task heads.

\begin{figure*}[t]

\begin{center}
\includegraphics[width=0.95\linewidth]{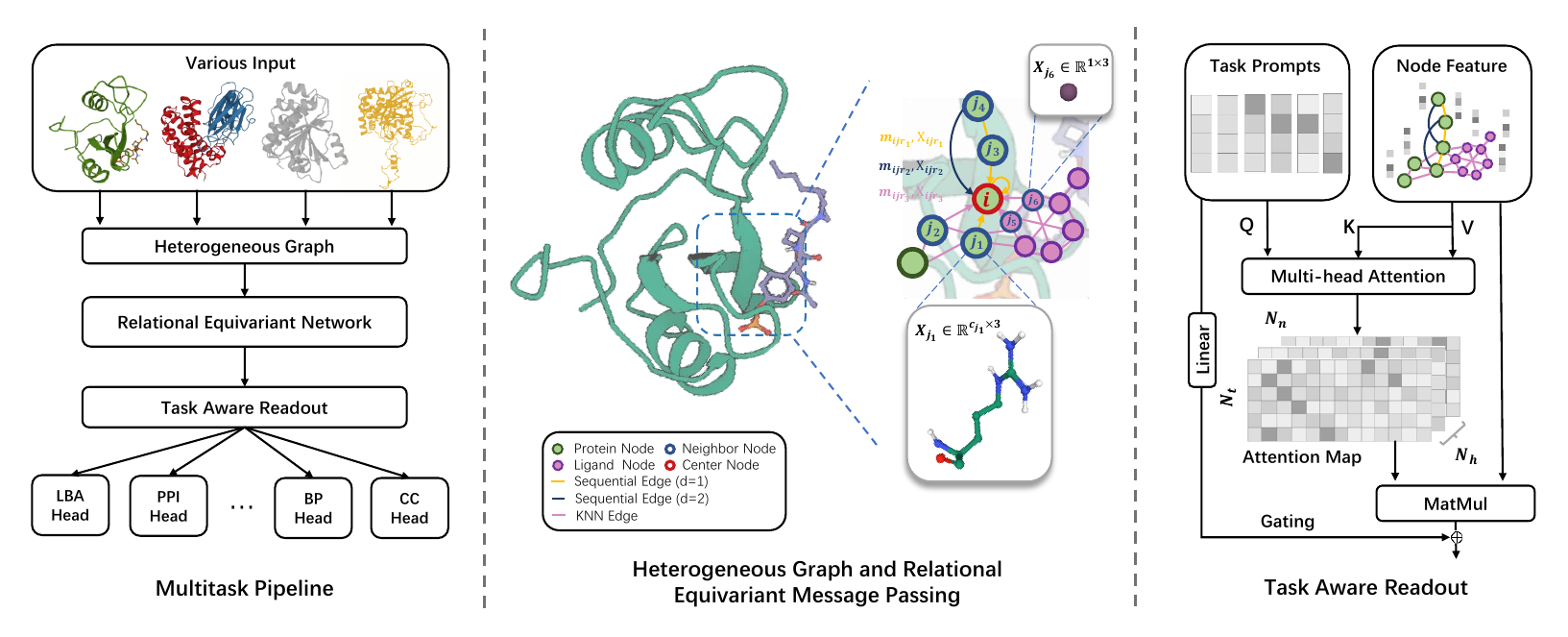}
\end{center}

\caption{\textbf{Overview of our pipeline}. Left: HeMeNet takes two-instance complexes or a single chain as input and predicts complex-level affinity and chain-level properties simultaneously. Middle: An example of the heterogeneous graph and the relational equivariant message passing. We only annotate a small part of our multi-channel full-atom graph for simplicity. Each edge is bidirectional, and we only mark the incoming edge arrow and self-loop for the center node. Right: Task-aware readout module. We take a task prompt as the query for each task, generating attention maps for all the nodes to get a multi-level readout for different downstream tasks.}
\label{fig2}
\end{figure*}

\paragraph{Heterogeneous Multi-channel Equivariant Message Passing} Inspired by dyMEAN~\cite{kong2023endtoend}, we leverage a multi-channel coordinate matrix with dynamic size to record the geometric information of a node in an input graph. Moreover, we extend the setting to heterogeneous message passing along multiple types of edges to capture rich relationships between nodes. We denote the node feature and coordinates as $(\vh_i^{(l)}, \vec{\mX}_i^{(l)})$ in the 
$l$-th layer. The message passing is calculated as:
\begin{align}
    \vm_{ijr} &= \phi_m(\vh_i^{(l)}, \vh_j^{(l)}, \frac{T_R(\displaystyle \vec{\mX}_i^{(l)}, \displaystyle \vec{\mX}_j^{(l)})}{||T_R(\displaystyle \vec{\mX}_i^{(l)}, \displaystyle \vec{\mX}_j^{(l)})||_F +\epsilon}, \ve_r), \\
    \displaystyle \vec{\mM}_{ijr} &= T_S(\displaystyle \vec{\mX}_i^{(l)} - \frac{1}{c_j}\sum_{k=1}^{c_j}\displaystyle \vec{\mX}_{j}^{(l)}(:, k), \phi_{x}(\vm_{ijr})),
 \label{E1}
\end{align}
where,  $\vm_{ijr}$ and $\vec{\mM}_{ijr}$ are separately the invariant and equivariant messages from node $j$ to $i$ along the $r$-th edge; $\ve_r$ is the edge  embedding feature; 
$\phi_m, \phi_x$ are Multi-Layer Perceptrons (MLPs)~\cite{gardner1998artificial} with one hidden layer; $||\cdot||_F$ computes the Frobenius norm, $T_R$ and $T_S$ are the adaptive multichannel geometric relation extractor and geometric message scaler, in order to deal with the issue incurred by the varying shape of $\vec{\mX}_i^{(l)}$ and $\vec{\mX}_j^{(l)}$ since the number of atoms could be different for different nodes. With the calculated messages, the node representation is updated by:
\begin{align}
   \vh_i^{(l+1)} &= \vh_i^{(l)} + \sigma(\mathrm{BN}(\phi_h(\sum_{r \in R} \mW_r\sum_{j \in \gN_r(i)}\vm_{ijr}))), \\ 
    \displaystyle \vec{\mX}_i^{(l+1)} &= \displaystyle \vec{\mX}_i^{(l)}
 + \frac{1}{\sum_{r \in R}|\mathcal{N}_r(i)|}\sum_{r \in R}\sum_{j \in \mathcal{N}_r(i)}w_r \displaystyle \vec{\mM}_{ijr},
 \label{E1}
\end{align}

where, $\mW_r, w_r$ are a learnable matrix and a learnable scalar to project invariant an equivariant messages, respectively, for the $r$-th kind of edge; $\gN_r(i)$ denotes the neighbor nodes of $i$ regarding the $r$-th kind of edges; $\phi_h$ is an MLP, $\mathrm{BN}$ is the batch normalization operation, and $\sigma$ is an activation function. During the message-passing process, our model gathers information from different relations for $\vh_i$ and $\vec{\mX}_i$, ensuring the E(3) equivariance. For further details of the components in our model, please refer to Appendix~\ref{appendix_b}.

\textbf{Task-Aware Readout}\quad After $L$ layers of the above relational message passing, we attain a set of  E(3) invariant node features $\mH^{(L)}\in\R^{n\times d_L}$, where $n$ is the number of nodes and $d_L$ is the feature dimension. To correlate the task-specific information with each node feature, we propose a task-aware readout function. We compute the attention weights between each node and each task-specific query, and then readout all weighted node features into dual-level representations: graph level or chain level for each task. As shown in Figure~\ref{fig2}, the task-aware readout module is formulated as:
\begin{align}
    \displaystyle \bm{\alpha}_{t} &= \mathrm{Softmax}(\frac{\displaystyle\displaystyle \mK \vq_{t}}{\sqrt{d_L}}),\\
   \displaystyle \vf_t &= \mathrm{FFN} (\displaystyle \bm{\alpha}_t\mV +  \mathrm{Linear}(\displaystyle \vq_t)),
 \label{E5}
\end{align}
where, $\bm{\alpha}_t\in[0,1]^{n}$ defines the attention values for task $t$; $\displaystyle \vq_t \in \mathbb{R}^{d_L}$ is the learnable query for task $t$; 
$\displaystyle \mK=\displaystyle \mH\displaystyle \mW_K\in\R^{n\times d_L}$ and $\displaystyle \mV=\displaystyle \mH\displaystyle \mW_V\in\R^{n\times d_L}$ are the key and value matrices, respectively; $\mathrm{FFN}$ is the feed-forward network containing layer normalization and linear layers;  $\mathrm{Linear}(\displaystyle \vq_t) = \displaystyle \mW_Q \displaystyle \vq_t + \displaystyle \vb$ is used before the shortcut addition. In our implementation, we apply the multi-head attention strategy by defining multiple queries for each task. Although we compute the attention by only using the invariant features $\mH^{(L)}$, it indeed has involved the geometric information from the 3D coordinates during the previous $L$-layer message passing. 

\textbf{Multiple Task Heads} \quad We feed the above task-specific feature $\displaystyle \vf_i$ into different task heads implemented by MLPs, resulting in a prediction list $ (\displaystyle \vp_{\mathrm{lba}}, \vp_{\mathrm{ppa}}, \vp_{\mathrm{ec}},\vp_{\mathrm{mf}}, \vp_{\mathrm{bp}}, \vp_{\mathrm{cc}})$.  For regression tasks (including LBA and PPA), we use the Mean Square Error (MSE) loss $\mathcal{L}_{\mathrm{MSE}}$. For classification tasks (including EC, GO-MF, GO-BP and GO-CC), we use the Binary Cross Entropy (BCE) loss $\mathcal{L}_{\mathrm{BCE}}$.
The training loss is formulated as:
\begin{align}
    \mathcal{L} &= \mathcal{L}_{\mathrm{reg}} + \lambda \mathcal{L}_{\mathrm{cls}},
    \label{E2}
\end{align}
where $\mathcal{L}_{\mathrm{reg}} = \displaystyle \1_\mathrm{lba}\mathcal{L}_{\mathrm{MSE}}(\displaystyle \vp_{\mathrm{lba}}) +  \displaystyle \1_\mathrm{ppa}\mathcal{L}_{\mathrm{MSE}}(\displaystyle \vp_{\mathrm{ppa}}), \mathcal{L}_{\mathrm{cls}} = \displaystyle \1_\mathrm{ec}\mathcal{L}_{\mathrm{BCE}}(\displaystyle \vp_{\mathrm{ec}}) + \displaystyle \1_\mathrm{mf}\mathcal{L}_{\mathrm{BCE}}(\displaystyle \vp_{\mathrm{mf}}) + \displaystyle \1_\mathrm{bp}\mathcal{L}_{\mathrm{BCE}}(\displaystyle \vp_{\mathrm{bp}}) + \displaystyle \1_\mathrm{cc}\mathcal{L}_{\mathrm{BCE}}(\displaystyle \vp_{\mathrm{cc}})$, $\lambda$ is a hyper-parameter to balance the trade-off of the losses. 
To allow training on the partially labeled sample, if the label of the task $*$ exists, then $\displaystyle \1_\mathrm{*} = \lambda_\mathrm{*}$, otherwise $\displaystyle \1_\mathrm{*} = 0$. In addition, we adopt a balanced sampling strategy to ensure that each sampled mini-batch should contain at least one sample from each task, which further accelerates the training convergence.

\section{Experiments}
In this section, we will first introduce the experimental setup in Section~\ref{sec:setup}.
In Section~\ref{sec:results}, we evaluate our model on the proposed dataset \datasetname{} for affinity and property prediction in both single-task and multi-task settings and compare it with other baseline models.
In Section~\ref{sec:readout}, we experiment with different readout strategies and compare their performance on property prediction tasks.
In Section~\ref{sec:ablation}, we perform ablation experiments on different modules.

\subsection{Experimental Setup}
\label{sec:setup}
\begin{table*}[t]
\setlength{\belowcaptionskip}{0.3cm}

\centering
\footnotesize
\begin{tabular}{cccccccccc}
\Xhline{1pt}
\multicolumn{1}{c}{}  & \multirow{2}{*}{Method} & \multicolumn{2}{c}{LBA} & \multicolumn{2}{c}{PPA} & \multicolumn{1}{c}{\multirow{2}{*}{EC$\uparrow$}} & \multicolumn{3}{c}{GO}                                           \\ \cmidrule(lr){3-4} \cmidrule(lr){5-6} \cmidrule(lr){8-10} 
\multicolumn{2}{c}{}  & \multicolumn{1}{c}{RMSE$\downarrow$} & \multicolumn{1}{c}{MAE$\downarrow$} & \multicolumn{1}{c}{RMSE$\downarrow$} & \multicolumn{1}{c}{MAE$\downarrow$} & \multicolumn{1}{c}{}                    & \multicolumn{1}{c}{MF$\uparrow$} & \multicolumn{1}{c}{BP$\uparrow$} & CC$\uparrow$           \\ \midrule
\multicolumn{1}{c}{\multirow{9}{*}{\rotatebox{90}{\textbf{Single-task}}}}
& GCN \cite{kipf2017semisupervised} & 2.193 & 1.721 & 7.840 & 7.738 & 0.022 & 0.207 & 0.254 & 0.367 \\

\multicolumn{1}{c}{} & GAT \cite{veličković2018graph} & 2.301 & 1.838 & 7.820 & 7.720 & 0.018 & 0.223 & 0.249 & 0.354 \\
\multicolumn{1}{c}{} & SchNet \cite{schutt2017schnet} & 2.162 & 1.692 & 7.839 & 7.729 & 0.097 & 0.311 & 0.281 & 0.431 \\
\multicolumn{1}{c}{} & GearNet* \cite{zhang2023protein} & 1.957 & 1.542 & 2.004 & 1.279 & 0.716 & 0.677 & 0.252 & 0.438 \\ 
\multicolumn{1}{c}{} & GearNet-fullatom \cite{zhang2023protein} & 2.178 & 1.716 & 2.753 & 2.709 & 0.046 & 0.212 & 0.229 & 0.471 \\ 
\cmidrule{2-10}

\multicolumn{1}{c}{} & EGNN \cite{EGNN}                & 2.282 & 1.849 & 4.854 & 4.756 & 0.039 & 0.206 & 0.253 & 0.357 \\
\multicolumn{1}{c}{} & GVP \cite{jing2021learning}     & 2.281  & 1.789   &  5.280 & 5.267 &  0.020 & 0.204 &    0.244   &    0.454   \\
\multicolumn{1}{c}{} & dyMEAN \cite{kong2023endtoend}  & 2.410 & 1.987 & 7.309 & 7.182 & 0.115 & 0.436 & 0.292 & \underline{0.477}\\
\multicolumn{1}{c}{} & HemeNet (Ours)                    & 1.912 & 1.490 & 6.031 & 5.891 & \textbf{0.863} & \textbf{0.778} & \textbf{0.404} & \textbf{0.544} \\ \cmidrule{1-10}
\multicolumn{1}{c}{\multirow{9}{*}{\rotatebox{90}{\textbf{Multi-task}}}} 

& SchNet \cite{schutt2017schnet} & 1.763 & 1.447 & 1.216 & 1.120 & 0.093 & 0.192 & 0.264 & 0.402 \\
\multicolumn{1}{c}{} & GearNet* \cite{zhang2023protein} & 2.193 & 1.863 & 1.275 & 1.035 & 0.187 & 0.203 & 0.261 & 0.379 \\
\multicolumn{1}{c}{} & GearNet-fullatom \cite{zhang2023protein} & 1.839 & 1.350 & 1.821 & 1.491 & 0.047 & 0.155 & 0.258 & 0.443 \\
\multicolumn{1}{c}{} & CDConv \cite{fan2022continuous} & \textbf{1.579} & \underline{1.352} & 2.386 & 1.822 & 0.324 & 0.246 & 0.241 & 0.424 \\


\cmidrule{2-10}
\multicolumn{1}{c}{} & EGNN \cite{EGNN}                & 1.777 & 1.441 & 0.999 & 0.821 & 0.048 & 0.169 & 0.244 & 0.352 \\
\multicolumn{1}{c}{} & GVP \cite{jing2021learning}     & 1.870  &   1.572    &   \underline{0.906}   &  \underline{0.758}    &   0.018    &   0.168    &   0.246   &   0.360     \\
\multicolumn{1}{c}{} & dyMEAN \cite{kong2023endtoend}  & 1.777 & 1.446 & 1.725 & 1.523 & 0.038 & 0.164 & 0.263 & 0.449 \\
\multicolumn{1}{c}{}  & HeMeNet* (Ours)  &  1.799 &   1.420    &   \textbf{0.861} &    \textbf{0.719}   &   0.630    &    0.595   &   0.279    &   0.426    \\
\multicolumn{1}{c}{} & HeMeNet (Ours)                    & \underline{1.730} & \textbf{1.335} & 1.087 & 0.912 & \underline{0.810} & \underline{0.727} & \underline{0.379} & 0.436 \\ 

\cmidrule{1-10}\morecmidrules\cmidrule{1-10}
\multicolumn{1}{c}{} & GPT4-turbo-1106 (ST) \cite{achiam2023gpt} & 2.347 & 1.780 & \textbf{1.654} & 1.343 &- &- & - & - \\
\multicolumn{1}{c}{} & ESM2 (MT) \cite{lin2023evolutionary} & 2.009 & 1.334 & 1.692 & \textbf{1.333} & 0.917 & 0.764 & 0.389 & 0.533 \\
\multicolumn{1}{c}{} & ESM2-HeMeNet (MT) & \textbf{1.867} & \textbf{1.661} & 1.846 & 1.418 & \textbf{0.921} & \textbf{0.796} & \textbf{0.455} & \textbf{0.567} \\

\Xhline{1pt}

\end{tabular}

\begin{tablenotes}
        \footnotesize
        \item * represents trained under the alpha-Carbon atom only setting. ST, MT is the abbreviation of single-task and multi-task, respectively.
      \end{tablenotes}
\caption{The mean result for three runs on the full-label test set. We select representative invariant and equivariant models for affinity prediction and property prediction. The upper half reports the results for the single-task setting, and the lower half reports the results for the multi-task setting. The best results are marked in bold and the second best results are underlined. In the multi-task setting, we train the models with the same size compared to their corresponding single-task models.}
\label{Table main}
\end{table*}

\textbf{Task settings} \quad
We compare the performance of HeMeNet with other models under single-task and multi-task settings using the same validation and test sets.
For single-task training, models are trained on samples with labels of the corresponding task.
We also remove the task-aware readout of our model for a fair comparison.
For multi-task training, the models are trained on all partially labeled training samples.

We use a balanced sampler for each batch to sample at least one complex from LBA and one complex from PPA.
We include samples with up to 15,000 atoms for training and evaluation.
All models are trained for 30 epochs on 4 NVIDIA A100 GPUs. More details on experimental settings can be found in Appendix~\ref{appendix_c}

\textbf{Baselines}\quad
We compared our model with nine representative baselines. 
\textbf{GCN}~\cite{kipf2017semisupervised} aggregates information weighted by the degree of nodes.
\textbf{GAT}~\cite{veličković2018graph} utilizes an attention mechanism for message passing. 
\textbf{Schnet}~\cite{schutt2017schnet} is an invariant network with continuous filter convolution on the 3D molecular graph.
\textbf{GearNet}~\cite{zhang2023protein} designs a relational message-passing network to capture information on protein function tasks. Its variant \textbf{Gearnet-fullatom}~\cite{zhang2023physics} utilizes the model to the full-atom setting. \textbf{CDConv}~\cite{fan2022continuous} models the geometric sequence with a continuous-discrete convolution.
Besides the previous invariant models, we also compare our method with equivariant models.
\textbf{EGNN}~\cite{EGNN} is a lightweight but effective E(n) equivariant graph neural network.
\textbf{GVP}~\cite{jing2021learning} designs an equivariant geometric vector perceptron for protein representation.
\textbf{dyMEAN}~\cite{kong2023endtoend} is an equivariant model for antibody design; it takes a dynamic multichannel equivariant function for full-atom coordinates. \textbf{ESM2}~\cite{lin2023evolutionary} is a general purpose protein language model. We also conducted three-shot prompting to test GPT-4~\cite{achiam2023gpt} on the two affinity prediction tasks, see Appendix~\ref{appendix_d} for the prompt details.

\textbf{Evaluation}\quad
For LBA and PPA tasks, we employ the commonly used Root Mean Square Error (RMSE) and Mean Average Error (MAE) as the evaluation metrics~\cite{Atom3d}.
For EC and GO tasks, we use maximum F-score (Fmax) following ~\cite{zhang2023protein}.
Each experiment is independently run three times with different random seeds.

\subsection{Results on \datasetname{}}
\label{sec:results}

We conduct experiments under both single-task and multi-task settings.
The mean results of three runs are reported in Table \ref{Table main}. 
According to the results, we draw conclusions summarized in the subsequent paragraphs.

\textbf{Our model outperforms the baselines on most of the tasks under both settings.} \quad
Under the single-task setting, our model surpasses other models in five of the six tasks.
Under the multi-task setting, our model surpasses other models in four of the six tasks, with the remaining two tasks reaching second and third place, respectively. We also compared our model with GPT4 and ESM2 (with multi-task head finetuning). Our method outperforms GPT4 in both LBA and PPA. ESM2 performs well on four property prediction tasks, and combining ESM2 and HeMeNet can further improve ESM2's overall performance with geometric information.
Notably, under the single-task setting, only the models with a heterogeneous message passing (GearNet and ours) can perform well on all of the four property prediction tasks.
Under the multi-task setting, our full-atom model, benefiting from joint learning, shows superior results on different tasks, and there are two main interesting observations discussed next.

\textbf{Our model benefits from the multi-task setting, especially on LBA and PPA.} \quad
We observe that almost all models improve their performance on LBA and PPA tasks under the multi-task setting.
In particular, our model significantly improves the PPA RMSE from 6.031 to 1.087 by utilizing a training set that is more than ten times larger (2587 for PPA single-task and 30904 for our multi-task training set). We also train our model with alpha C atom (HeMeNet*) as input, resulting in a best PPA RMSE of 0.861. To further understand the internal transfer of information within the tasks, we choose several different ratios to include different amounts of samples from the PPA tasks and keep the samples from other tasks unchanged. As shown in Figure \ref{fig4}, our model performs better as the training samples of PPA increase. And it performs much better than its single-task counterpart, even with a small amount of training samples.
These results demonstrate that the model can handle challenging tasks (complex-level affinity prediction) better when more structural information is available (e.g., single-chain structures and their labels).

\textbf{Our model performs harmonious multi-task training on property prediction tasks.} \quad
We observe that when switching from the single-task to multi-task setting, baseline models experience performance degradation to some extent across the four property prediction tasks.

This is probably because of task interference among diverse tasks, and combining different tasks for training without careful adaption can harm performance.

With the guidance of our task-aware readout module, our model is able to learn from multiple tasks in a task-harmonious way, while achieving performance on the property prediction tasks comparable to their single-task counterparts with the same parameter size.


\begin{table}[htbp]
 \begin{tabular}{ccccccccc}
        \toprule
        Method                & EC$\uparrow$     & GO-MF$\uparrow$ & GO-BP$\uparrow$ & GO-CC$\uparrow$ \\ 
        \cmidrule(lr){1-5}
        \textbf{Gearnet$_s$}  & 0.187 & 0.203 & 0.261 & 0.379 \\
        \textbf{Gearnet$_w$}  & 0.066 & 0.164 & 0.271 & 0.414   \\
        \textbf{Gearnet$_t$}  & 0.421 & 0.310 & 0.287 & 0.403 \\
        \textbf{HeMeNet$_s$}  & 0.722 & 0.558 & 0.302 & 0.413    \\
        \textbf{HeMeNet$_w$}  & 0.325 & 0.312 & 0.276 & \textbf{0.440}     \\
        \textbf{HeMeNet$_t$}  & \textbf{0.810} & \textbf{0.727} & \textbf{0.379} & 0.436 \\
        \bottomrule
        \end{tabular}
        \caption{Comparison of different readout functions for multi-task learning. $s$, $w$, and $t$ represent sum, weighted feature and task-aware readout, respectively.}
        \label{Table 2}
\end{table}

\subsection{Comparison of Different Readout Methods}
\label{sec:readout}

To verify the effectiveness of our task-aware readout module, we take HeMeNet and Gearnet as the backbone and compare the task-aware readout method with two commonly used readout functions: sum readout and task-prompt weighted node feature~\cite{liu2023graphprompt}.
The results are presented in Table~\ref{Table 2}. We can conclude with the following observations:
\begin{enumerate*}[1)]
    \item Our proposed task-aware readout model injects task-related information using an attention mechanism, leading to overall improvements for various tasks, especially on the Enzyme Commission task.
    \item Simply element-wise multiplication of the task prompt feature with all the nodes fails to provide sufficient guidance to learning across all tasks.
\end{enumerate*}


\begin{figure}[htbp]
\footnotesize
    \centering
	\begin{minipage}{0.45\linewidth}
		\centering

		\setlength{\abovecaptionskip}{0.28cm}
		\includegraphics[width=\linewidth]{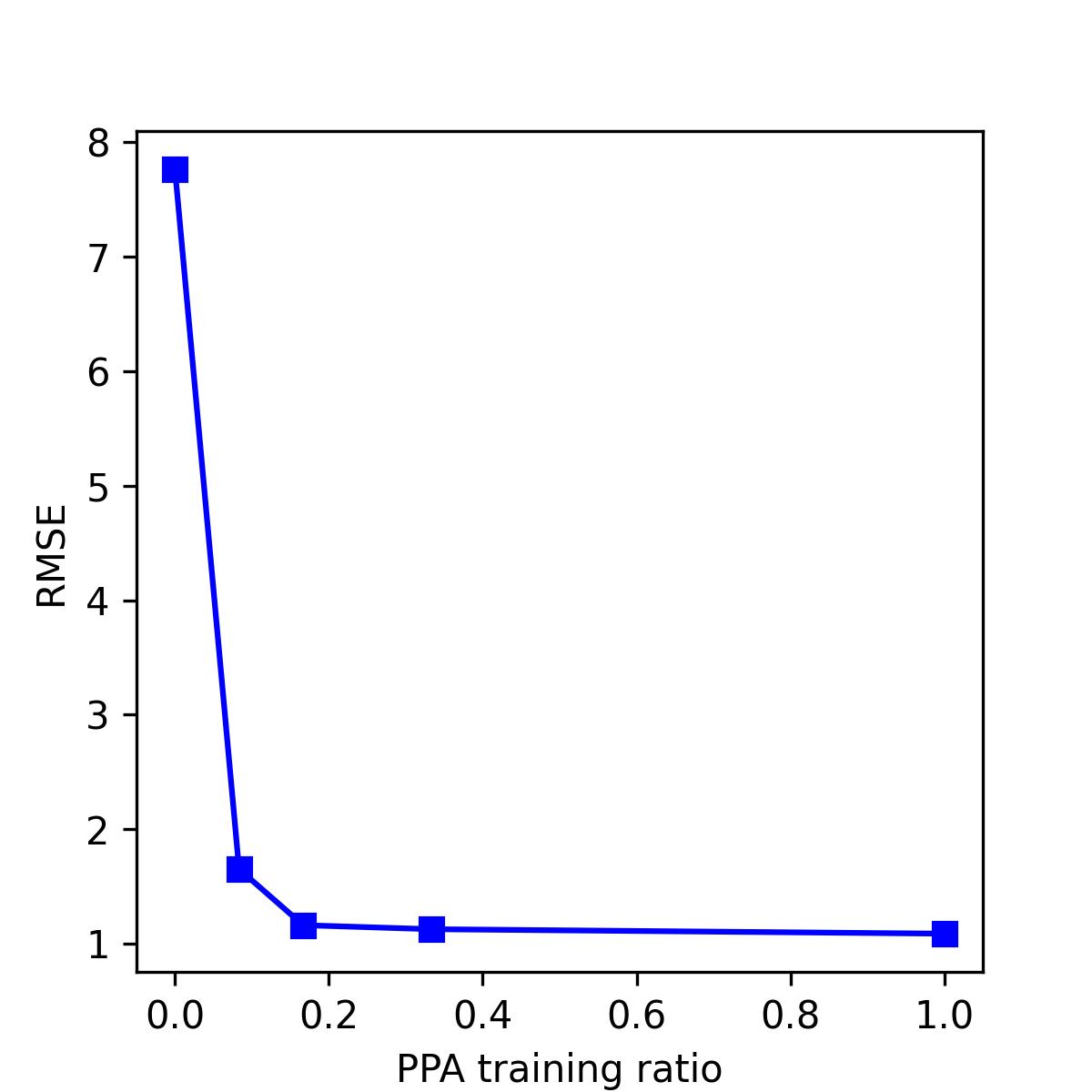}
		\caption{PPA performance}
		\label{fig4}
	\end{minipage}
	\hfill
	\begin{minipage}{0.48\linewidth}
        \centering
		\setlength{\abovecaptionskip}{0.55cm}
  		\vspace{0.2cm}
		\includegraphics[width=\linewidth]{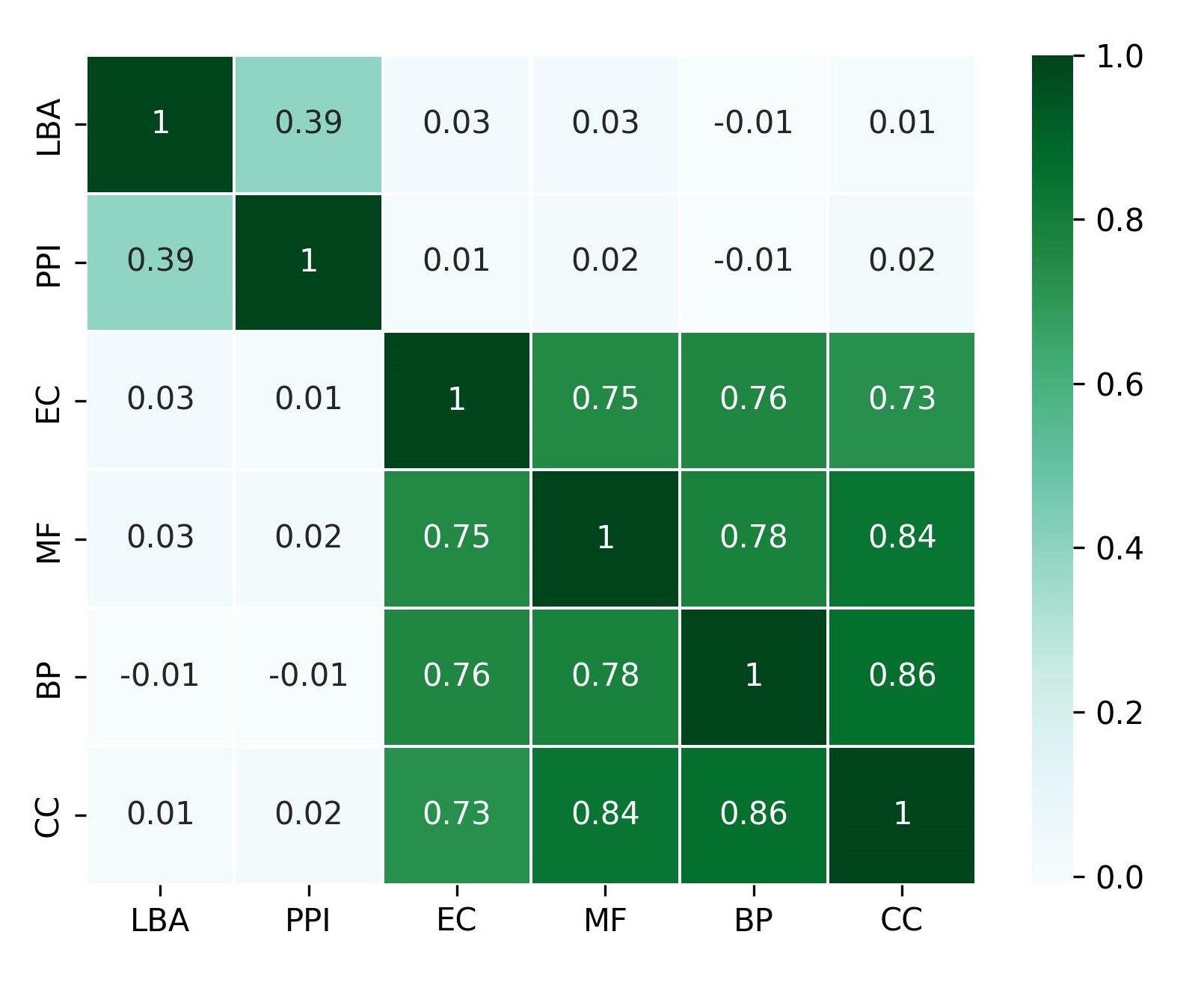}
		\caption{Prompt correlation}
		\label{fig5}
		
	    \end{minipage}
\end{figure}

To further investigate the relationship between multiple tasks, we calculate Pearson's correlation between prompts. 
As shown in Figure~\ref{fig5}, the correlations between tasks within the same category (e.g. EC and MF) are high, while the correlations between tasks from different categories (e.g. LBA and BP) are low.
A high correlation between prompts indicates similar attention queries, leading to similar readout functions.
This suggests that with the task-aware guidance, the model employs similar readout strategies for tasks from the same category and divergent strategies from tasks from different categories.

\begin{table}[htbp]
\setlength{\belowcaptionskip}{0.3cm}

\centering
\footnotesize
\begin{tabular}{lcccccc}
\Xhline{1pt}
\multirow{2}{*}{Method}  & \multicolumn{1}{c}{\multirow{2}{*}{LBA$\downarrow$}}   & \multicolumn{1}{c}{\multirow{2}{*}{PPA$\downarrow$}}  & \multicolumn{1}{c}{\multirow{2}{*}{EC$\uparrow$}} & \multicolumn{3}{c}{GO}   \\  \cmidrule(lr){5-7} 
& \multicolumn{1}{c}{} & \multicolumn{1}{c}{} & \multicolumn{1}{c}{} & \multicolumn{1}{c}{MF$\uparrow$} & \multicolumn{1}{c}{BP$\uparrow$} & CC$\uparrow$           \\ \midrule
HeMeNet & \textbf{1.730} & \underline{1.087} & \textbf{0.810} & \textbf{0.727} & \textbf{0.379} & \textbf{0.436} \\
- TAR & 1.905 & 1.970 & 0.722 & 0.558 & 0.302 & 0.413 \\ 
- $e_r, W_r, w_r$ &  1.790  & 1.446 & 0.547 & 0.663 & 0.359 & 0.391 \\
- full-atom   &  1.799  &   \textbf{0.861}  &   0.630    &    0.595   &   0.279    &   0.426    \\

\Xhline{1pt}
\end{tabular}

\caption{Ablation study for different components in HeMeNet.}
\label{Table 3}

\end{table}

\vspace{-0.5cm}
\subsection{Ablation Study}
\label{sec:ablation}
We perform ablation experiments to evaluate the necessity of different components, including the task-aware readout, the relational message passing mechanism, and the full-atom geometry. Specifically, the ablation of TAR replaces the task-aware readout module with a sum readout. For $e_r, W_r$ and $w_r$, we remove different types of edges and the relational message passing weights. For full-atom ablation, we represent the coordinates of residues by their alpha C atoms.

We present the results for ablation studies in Table~\ref{Table 3}, the observations are as follows: 
\begin{enumerate*}[1)]
    \item Without our task-aware readout strategy, significant performance degradation are observed in all tasks, indicating that the tasks can hinder each other without appropriate guidance.
    \item Without the heterogeneous graph and the relational message passing, our model's performance drops on property prediction tasks, especially the Enzyme Commission number prediction.
    \item Removing the full-atom geometry decreases the performance in multiple tasks. However, it improves our model's performance in PPA. Similar to the explanation in Section~\ref{sec:results}, we suppose that the large number of atoms in the full-atom protein-protein complex introduces excessive noise to prediction compared with input with alpha-Carbon atoms only.
\end{enumerate*}

\section{Conclusion}

In this paper, we alleviate the problem of sparse data in structured protein datasets by a multi-task setting. We construct a standard multi-task benchmark \datasetname{}, consisting of 6 representative tasks integrated from 4 public datasets for joint learning. We propose a novel network called HeMeNet to address multiple tasks in protein 3D learning, with a novel heterogeneous equivariant full-atom encoder and a task-aware readout module. Comprehensive experiments demonstrate our model's performance on the affinity and property prediction tasks. Our work brings insights for utilizing different structural datasets to train a more powerful generalist model in future research.

\bibliographystyle{named}
\bibliography{ijcai24}
\newpage
\appendix

\section{Data}  \label{appendix_a}
\subsection{Dataset sources}
\subsubsection{Enzyme Commission and Gene Ontology}
We adopt the data set from \cite{gligorijevic2021structure}. This data set contains 19201 PDB chains from 538 EC numbers, selected from the third and fourth levels of EC tree. The GO terms with at least 50 and not more than 5,000 training examples are selected. The number of classes in each branch of GO (MF, BP, CC) is 490, 1944 and 321, respectively. This data set consolidates 36641 PDB chains with their GO terms label. We obtain the structure of PDB chains from TorchDrug \cite{zhu2022torchdrug}. 
\subsubsection{LBA and PPA} We adopt the data set from PDBbind \cite{PDBbind2004} (version 2020). This dataset contains 5316 protein-ligand complexes in the refined set and 2852 protein-protein complexes with known binding data (in the form of $K_d, \ K_i, \ IC_{50}$ values) manually collected from the original references \cite{PDBbind2004}. We obtain the structure of the complexes from PDBbind. The PDBbind dataset can be downloaded from \url{http://www.pdbbind.org.cn}.
\subsection{\datasetname{} statstics}

As described in section \ref{sec:dataconstruction}, we yield 1327 fully-labeled complexes.Specifically, we employ MMSeq2 to cluster all the chains in \datasetname{} with an alignment coverage $>30\%$
and sequence identity of the aligned fragment $>90\%$, leading to 33704 chain-level clusters. Then, we merge the clusters that contain the chains belonging to the same complex and finally get 30034 clusters. For the fully-sampled complexes, we randomly split them into the training, validation and test sets, with the number of complexes as 328, 530, and 469, respectively. For the partially labeled samples, we only retain those samples located in clusters different from the above test complexes and add them into the training set, resulting in an augmented training set with a total of 31252 samples.
After the procedure of labeling and splitting, we get a new dataset named \datasetname{}.
Table \ref{Table a1} shows the detailed statistics of \datasetname{} in different tasks. Note that the sample number for multi-task is slightly different from that in Table \ref{Table a0} since we removed samples with atom numbers greater than 15,000.

\begin{table*}[htbp]
\centering
\setlength{\belowcaptionskip}{0.3cm}
\caption{Dataset split. The fully-labeled data are randomly divided into the train, validation and test sets. Partially labeled samples located in the clusters different from the above test complexes are retained and added to the training set.}
\begin{tabular}{cccc}
\Xhline{1pt}
Clusters after merged & Train set size & Validation set size & Test set size \\
\midrule
30034 & 31252 & 530 & 469\\
\Xhline{1pt}
\end{tabular}
\label{Table a0}
\end{table*}

\begin{table*}[htbp]
\centering
\setlength{\belowcaptionskip}{0.3cm}
\caption{Dataset details for different tasks. We summarize the number of samples that contains a specific task's annotation in Protein-MT.}
\begin{tabular}{cccccccc}
\Xhline{1pt}
Task Name & \# \makecell[c]{Train \\ samples} & \# \makecell[c]{Validation \\ samples} & \# \makecell[c]{Test \\ samples}  & \makecell[c]{Training label\\ missing ratios} & \makecell[c]{Average \\seq length}& \# Average atoms & Chains\\
\midrule
multi-task & 30904 & 516 & 467 & 99.9\% &325.67 & 2537.68 & Mixed \\
LBA & 3247 & 493 & 452 & 89.5\% & 425.29 & 3337.21 & Multi\\
PPA & 1976 & 23 & 15 & 93.6\% & 846.53 & 6604.35 & Multi\\
EC & 15025 & 516 & 467 & 51.4\% & 331.68 & 2585.91 & Single\\
GO-MF & 22997 & 516 & 467 & 25.6\% & 311.18 & 2420.01 & Single\\
GO-BP & 21626 & 516 & 467 & 30.0\% & 305.90 & 2382.08 & Single\\
GO-CC & 10543 & 516 & 467 & 65.9\% & 302.47 & 2346.91 & Single\\
\Xhline{1pt}
\end{tabular}
\label{Table a1}
\end{table*}

\section{Details for HeMeNet components} \label{appendix_b}
In this section, we mainly introduce $T_R$ and $T_S$ from \cite{kong2023endtoend} and our modification for the heterogeneous graph. 

\textbf{The Geometric Relation Extractor $T_R$} can deal with coordinate sets with different channels. Given $\displaystyle \mX_i \in \mathbb{R}^{3\times c_i}$ and $\displaystyle \mX_j \in \mathbb{R}^{3\times c_j}$, we can compute the channel-wise distance for each coordinate pairs: $D_{ij}(p,q) = ||\displaystyle \mX_{i}(:, p) - \displaystyle \mX_{j}(:,q)||_{2}$. Different from \cite{kong2023endtoend}, we use two fixed binary vectors $\displaystyle \vw_i \in \mathbb{R}^{c_i\times 1}$ and $\displaystyle \vw_j \in \mathbb{R}^{c_j\times 1}$, when there is an element in the channel, its weight is set to 1, otherwise 0. We also adjusted the learnable attribute matrices $\displaystyle \mA_i \in \mathbb{R}^{c_i \times d}$ and $\displaystyle \mA_j \in \mathbb{R}^{c_j \times d}$ to be suitable to our input, assigning different element embedding for each channel. The final output $\displaystyle \mR_{ij} \in \mathbb{R}^ {d \times d}$  is given by:

\begin{equation}
    \displaystyle \mR_{ij} = \displaystyle \mA^T_i(\displaystyle \vw_i \displaystyle \vw_j^T \odot \displaystyle \mD_{ij}) \displaystyle \mA_j.
    \label{eq8}
\end{equation}

$\displaystyle \mR_{ij}$ keeps its shape awareness of $c_i$ and $c_j$.

\textbf{The Geometric Message Scaler $T_S$} aims to generate geometric information of vary coordinate set $\displaystyle \mX \in \mathbb{R}^{3 \times c}$ with the fixed length incoming message $s = \phi_x(\displaystyle \vm_{ij}) \in \mathbb{R}^C$, where $C=14$ is the max channel size of the common amino acids. Then, $T_S(\displaystyle \mX, \displaystyle \vs)$ is calculated by:

\begin{equation}
    \displaystyle \mX' = \displaystyle \mX \cdot \mathrm{diag}(s'),
    \label{eq9}
\end{equation}

where $s' \in \mathbb{R}^c$ is the average pooling of $\displaystyle \vs$ with a sliding window of size $C-c+1$ and stride 1, and $\mathrm{\cdot}$ is a diagonal matrix with the input vector $s$ as the diagonal elements.

\section{Implementation Details and hyperparameters} \label{appendix_c}
In this section, we introduce the implementation details of all baselines and our model. For all models, we concatenate the hidden output for the final output. For the multi-task setting, all the models except HeMeNet take the sum readout method. The feature after the readout function will be fed into six two-layer MLPs to make predictions for different prediction tasks. The input for models are full-atom with KNN edges, except for GearNet and HeMeNet.

\textbf{GCN \cite{kipf2017semisupervised}}, \textbf{GAT \cite{veličković2018graph}} and \textbf{SchNet \cite{schutt2017schnet}}. We take the implementation in PyTorch Geometric \cite{Fey/Lenssen/2019}, with a 3-layer design. For all the models, the hidden size is set to 256.

\textbf{GearNet \cite{zhang2023protein}}. We re-implement GearNet with reference to its official implementation, with a six-layer design. The hidden size is set to 512, and the cutoff value is 4.5 following the original settings. For the multi-task setting, we take the sum readout method. We use the alpha-Carbon atom only graph for GearNet as the input.

\textbf{EGNN \cite{EGNN}}. We re-implement EGNN with reference to its official implementation, with a 3-layer design. The hidden size is set to 256.

\textbf{GVP \cite{jing2021learning}}.We take the implementation in PyTorch Geometric \cite{Fey/Lenssen/2019}, with a 3-layer design. The hidden size is set to 128 following the original implementation.

\textbf{dyMEAN \cite{kong2023endtoend}}. We re-implement dyMEAN with reference to its official implementation, with a 6-layer design. The hidden size is set to 256.

\textbf{HeMeNet (ours)}. We take a 6-layer design for our model, and the hidden size is set to 256. We take our task-aware readout module to generate features for different tasks. We use the full-atom heterogeneous graph for HeMeNet as the input.

\section{More fine-grained task combinations}
\label{appendix_d}

In order to findout the effect of different tasks to LBA\&PPA tasks, we further train our model on different task combinations (LBA\&PPA+one property task). And the results can be seen in Table~\ref{Table combination}.

\begin{table}[htbp]
\centering
\setlength{\belowcaptionskip}{0.3cm}
\caption{Results on different task combinations.}
\begin{tabular}{ccccccc}
\Xhline{1pt}
Tasks & LBA & PPA & EC & MF & BP & CC\\
\midrule
LP\&EC & 1.758 & 1.064 & 0.628 & - & - & -\\
LP\&MF & 1.787 & 1.190 & - & 0.671 & - & -\\
LP\&BP & 1.820 & 1.089 & - & - & 0.374 & -\\
LP\&CC & 1.798 & 1.215 & - & - & - & 0.392 \\
All & \textbf{1.730} & \textbf{1.087} & \textbf{0.810} & \textbf{0.727} & \textbf{0.379} & \textbf{0.436} \\ 
\Xhline{1pt}
\end{tabular}
\label{Table combination}
\end{table}

As shown in the table, different property prediction tasks can improve the performance of LBA and PPA prediction, compared with the single-task settings. However, adding six tasks together will result in an overall better result. We suppose this is because the property prediction tasks shares a high correlation with each other, and each of them contains information from different aspects. Combining them will inprover the overall performance, benefiting our model by increasing the sample size and label diversity.
\section{Equivariance of HeMeNet} 
\label{appendix_e}

In this section, we will provide the equivariance of out HeMeNet encoder. Notice that the task-aware readout function only utilize the invariant feature with geometric information encoded for the downstream invariant tasks, including affinity prediction and property prediction. Therefore, we will prove the equivariance of the encoder and the invariance of the task-aware readout function.

\paragraph{Theorem 1. Equivariance of the HeMeNet encoder}

\textit{Given the input ($h_i$,$\displaystyle \mX_i$), we have the output ($\Tilde{h}_i$,$\Tilde{\displaystyle \mX_i}$) = $\mathrm{H}_e(h_i,\displaystyle \mX_i)$, where $\mathrm{H}_e$ is the abbreviation of HeMeNet encoder. $\mathrm{H}_e$ possesses the good property of E(3) equivariance. In other words, for any transformations $g \in E(3)$, we have ($\Tilde{h}_i$,$g \cdot \Tilde{\displaystyle \mX_i}$) = $\mathrm{H}_e(h_i,g \cdot \displaystyle \mX_i)$}

\paragraph{Lemma 1.} The geometric relation extractor is E(3) invariant. Specifically, $\forall \displaystyle \mX_i \in \mathbb{R}^{3 \times c_i}$, $\displaystyle \mX_j \in \mathbb{R}^{3 \times c_j}$, suppose $\displaystyle \mR_{ij} = \mathrm{T_R}(\displaystyle \mX_i, \displaystyle \mX_j)$, for any $g \in E(3)$, we have $\displaystyle \mR_{ij} = \mathrm{T_R}(g \cdot \displaystyle \mX_i, g \cdot \displaystyle \mX_j)$, where $g = \displaystyle \mQ x + t, Q \in O(3), t \in \mathbb{R}^3$.

\noindent \textit{Proof.} Since in the calculation of $\mathbf{R}_{ij}$, only $\mathbf{D}_{ij}$ is related to the input of $\mathbf{X}$ (as shown in Equation \ref{eq8}), the invariance of $\mathbf{R}_{ij}$ is equivalent of the invarinace of $\mathbf{D}_{ij}$. We now prove the invariance of $\mathbf{D}_{ij}$:

\begin{equation}
    \begin{split}
    \mathbf{D}_{ij} & = ||(Q \displaystyle \mX_i(:,p) + t) - (Q \displaystyle \mX_j(:,q) + t)||_2  \\
    & = ||Q (\displaystyle \mX_i(:,p) -  \displaystyle \mX_j(:,q))||_2  \\
    & = \sqrt{[\displaystyle \mX_i(:,p) - \displaystyle \mX_j(:,q)]^TQ^TQ[\displaystyle \mX_i(:,p) - \displaystyle \mX_j(:,q)]} \\
    & = \sqrt{[\displaystyle \mX_i(:,p) - \displaystyle \mX_j(:,q)]^T[\displaystyle \mX_i(:,p) - \displaystyle \mX_j(:,q)]} \\
    & = ||\displaystyle \mX_i(:,p) - \displaystyle \mX_j(:,q)||_2 
    \label{eq10}
    \end{split}
\end{equation}

The other terms in $\mathrm{T_R}$ ,such as $\displaystyle \mA_i$ and $\omega$, dose not change with respect to the transformation of $\displaystyle \mX_{ij}$. Therefore, $\mathbf{R}_{ij}$ is E(3) invariant. Similarly, because other parts in Equation \ref{E1} dose not change with respect to the input of $\mathbf{X}$. Therefore, $\displaystyle \mR_{ij} = \mathrm{T_R}(g \cdot \displaystyle \mX_i, g \cdot \displaystyle \mX_j)$. $\hfill\qed$

\paragraph{Lemma 2.} The geometric message scaler $\mathrm{T_S}$ is O(3) equivariant. Specifically, $\forall \displaystyle \mX \in \mathbb{R}^{3\times c}, s\in \mathbb{R}^C$, suppose $X' = \mathrm{T_S}(\displaystyle \mX, \displaystyle \vs)$, $\forall Q \in O(3)$, we have $Q\displaystyle \mX' = \mathrm{T_S}(Q\displaystyle \mX, \displaystyle \vs)$.

\noindent \textit{Proof.} We can derive as follows: 

\begin{equation}
    \begin{split}
    \mathrm{T_S}(Q\displaystyle \mX, \displaystyle \vs) &= Q\displaystyle \mX \cdot diag(s') \\
    &= Q(\displaystyle \mX \cdot diag(s'))\\
    &= Q\mathrm{T_S}({\displaystyle \mX, \displaystyle \vs}) \\
    &= Q\displaystyle \mX'
    \label{eq11}
    \end{split}
\end{equation}
Therefore, $Q\displaystyle \mX' = \mathrm{T_S}(Q\displaystyle \mX, \displaystyle \vs)$. $\hfill\qed$

With these lemmas, we can prove Theorem 1 now:

\noindent \textit{Proof.} Since $\displaystyle \vh_i, \displaystyle \vh_j, \displaystyle\ve_r$ will not change with respect to the transformation of $\displaystyle \mX$, $\forall g \in E(3)$, we can get:

\begin{equation}
    \begin{split}
    \displaystyle \vm_{ijr} &= \phi_m(\displaystyle \vh_i^{(l)}, \displaystyle \vh_j^{(l)}, \frac{\mathrm{T_R}(g\cdot\displaystyle \vec{\mX}_{i}^{(l)}, g\cdot\displaystyle \vec{\mX}_j^{(l)})}{||\mathrm{T_R}(g\cdot\displaystyle \vec{\mX}_{i}^{(l)}, g\cdot\displaystyle \vec{\mX}_j^{(l)})||_F+\epsilon}) \\
     &= \phi_m(\displaystyle \vh_i^{(l)}, \displaystyle \vh_j^{(l)}, \frac{\mathrm{T_R}(\displaystyle \vec{\mX}_{i}^{(l)}, \displaystyle \vec{\mX}_j^{(l)})}{||\mathrm{T_R}(\displaystyle \vec{\mX}_{i}^{(l)}, \displaystyle \vec{\mX}_j^{(l)})||_F+\epsilon}),
    \label{eq12}
    \end{split}
\end{equation}

and with the invariance of $\displaystyle \vm_{ijr}$, $\forall g \in E(3)$, that is, $\forall g\cdot \displaystyle \mX = Q\displaystyle \mX + \displaystyle \vt$, we can now derive the equivariance of $\displaystyle \vec{\mM}_{ijr}$:

\begin{equation}
    \begin{split}
    g\cdot \displaystyle \vec{\mM}_{ijr} &= \mathrm{T_S}(Q\displaystyle \vec{\mX}_i^{(l)}  + \displaystyle \vt- \frac{1}{c_j}\sum_{k=1}^{c_j} Q\displaystyle \vec{\mX}_j^{(l)}(:,k) \\
    &+ \displaystyle \vt, \phi_x(\displaystyle \vm_{ijr})) \\
   &= \mathrm{T_S}(Q\displaystyle \vec{\mX}_i^{(l)}  - \frac{1}{c_j}\sum_{k=1}^{c_j} Q\displaystyle \vec{\mX}_j^{(l)}(:,k), \phi_x(\displaystyle \vm_{ijr})) \\
   & = Q\mathrm{T_S}(\displaystyle \vec{\mX}_i^{(l)}  - \frac{1}{c_j}\sum_{k=1}^{c_j} \displaystyle \vec{\mX}_j^{(l)}(:,k), \phi_x(\displaystyle \vm_{ijr})),
    \label{eq13}
    \end{split}
\end{equation}

we can now prove the E(3) equivariance of our HeMeNet encoder $H_e$:
\begin{equation}
    \begin{split}
   g\cdot\vh_i^{(l+1)} &= \vh_i^{(l)} + \sigma(\mathrm{BN}(\phi_h(\sum_{r \in R} \mW_r\sum_{j \in \gN_r(i)}g\cdot\vm_{ijr}))) \\
   &= \vh_i^{(l)} + \sigma(\mathrm{BN}(\phi_h(\sum_{r \in R} \mW_r\sum_{j \in \gN_r(i)}\vm_{ijr}))) \\
   &= \vh_i^{(l+1)}, \\
    g\cdot\displaystyle \vec{\mX}_i^{(l+1)} &= g\cdot\displaystyle \vec{\mX}_i^{(l)}
 + \frac{1}{\sum_{r \in R}|\mathcal{N}_r(i)|}\sum_{r \in R}\sum_{j \in \mathcal{N}_r(i)}w_r g\cdot\displaystyle \vec{\mM}_{ijr} \\
  &=Q\vec{\mX}_i^{(l)} + \displaystyle\vt
 + \frac{1}{\sum_{r \in R}|\mathcal{N}_r(i)|}\sum_{r \in R}\sum_{j \in \mathcal{N}_r(i)}w_r Q\displaystyle \vec{\mM}_{ijr}\\
  &= Q[\displaystyle \vec{\mX}_i^{(l)}
 + \frac{1}{\sum_{r \in R}|\mathcal{N}_r(i)|}\sum_{r \in R}\sum_{j \in \mathcal{N}_r(i)}w_r \displaystyle \vec{\mM}_{ijr}] + \displaystyle \vt\\
 &= Q\displaystyle \vec{\mX}_i^{(l+1)} + \displaystyle \vt.
 \label{eq14}
\end{split}
\end{equation}

Therefore, $\forall g \in E(e)$, ($\Tilde{h}_i$,$g \cdot \Tilde{\displaystyle \mX_i}$) = $\mathrm{H}_e(h_i,g \cdot \displaystyle \mX_i)$. $\hfill \qed$

\paragraph{Theorem 2. Invariance of the Task-aware Readout}
\noindent\textit{Given the input $(\Tilde{h}, \Tilde{\displaystyle \mX})$, we have the output $(h', \displaystyle \mX') = \mathrm{TAR}(\Tilde{h}, \Tilde{\displaystyle \mX})$. TAR possesses the property of E(3) invariance. In other words, for any transformation $g\in E(3)$, we have $(h', \displaystyle \mX') = \mathrm{TAR}(\Tilde{h}, g\cdot\Tilde{\displaystyle \mX})$.}

\noindent \textit{Proof.} According to Equation \ref{E5}, the TAR only calculates over the invariant feature $\Tilde{h}$. The coordinates information $\Tilde{\displaystyle \mX}$ is identically set as the final output. Therefore, TAR is E(3) invariant. $\hfill \qed$

\section{GPT4 Prompt information} \label{appendix_f}

To test how GPT4 can be used to predict the binding affinity of protein-protein and protein-ligand, we provide the sequence of the receptor and the ligand (in the form of either SMILES or protein sequence)
. The system prompt is designed as follows:
\begin{tcolorbox}[
			colframe=black,
			width=9cm,
			arc=2mm, auto outer arc,
			title={System Prompt},breakable,]		
			System Prompt: You are a drug assistant and should be able to help with drug discovery tasks. Given the SMILES sequence of a drug and the FASTA sequence of a protein target, you need to calculate the binding affinity score. You can think step-by-step to get the answer and call any function you want. You should try your best to estimate the affinity with tools. The output should be a float number, which is the estimated affinity score without other words.
\end{tcolorbox}

We implemented a 3-shot in-context prompt as the input, we provide 3 smiles, protein sequences and affinity examples as follows:

\begin{tcolorbox}[
    colframe=black,
    width=9cm,
    arc=2mm, auto outer arc,
    title={User Prompt},breakable,]		
    Example 1:

    CC[C@H](C)[C@H](NC(=O)OC)C(=O)N1CCC[C@H] 1c1ncc(-c2ccc3cc(-c4ccc5[nH]c([C@@H]6CCCN6C(=O) [C@@H](NC(=O)OC)[C@@H](C)OC)nc5c4)ccc3c2)
     [nH]1,

    MDSIQAEEWYFGKITRRESERLLLNAENPRGTFLVR ESETTKGAYCLSVSDFDNAKGLNVKHYKIRKLDS GGFYITSRTQFNSLQQLVAYYSKHADGLCHRLTT VCP

    11.52

    Example 2:
    
    [H]C1:C([H]):C(S(=O)(=O)N([H])[H]):C([H]):C([H]):C: 1/N=N/N1C([H])([H])C([H])([H])C([H])([H])C([H])([H]) C([H])([H])C1([H])[H],
    
    HWGYGKHNGPEHWHKDFPIAKGERQSPVDIDTHT AKYDPSLKPLSVSYDQATSLRILNNGHAFNVE FDDSQDKAVLKGGPLDGTYRLIQFHFHWGSL DGQGSEHTVDKKKYAAELHLVHWNTKYGDFG KAVQQPDGLAVLGIFLKVGSAKPGLQKVVDV LDSIKTKGKSADFTNFDPRGLLPESLDYWTY PGSLTTPPLLECVTWIVLKEPISVSSEQVLKFRK LNFNGEGEPEELMVDNWRPAQPLKNRQIKASFK
    6.5

    Example 3:

    [H]/C1=C(C([H])([H])C(=O)N([H])C2:C([H]):C([H]) :C(S(=O)(=O)N([H])[H]):C([H]):C:2[H])C2:C([H]): C([H]):C(OC([H]) ([H])[H]):C([H]):C:2OC1=O,

    HWGYGKHNGPEHWHKDFPIAKGERQSPVDIDTHT AKYDPSLKPLSVSYDQATSLRILNNGHAFNVE FDDSQDKAVLKGGPLDGTYRLIQFHFHWGSLD GQGSEHTVDKKKYAAELHLVHWNTKYGDFGKA VQQPDGLAVLGIFLKVGSAKPGLQKVVDVLDS IKTKGKSADFTNFDPRGLLPESLDYWTYPGSL TTPPLLECVTWIVLKEPISVSSEQVLKFRKLN FNGEGEPEELMVDNWRPAQPLKNRQIKASFK\\

    For smiles, fasta in messages
    
    Test input:
    
    \{smiles\},
    
    \{fasta\},
\end{tcolorbox}

\section{Limitations and future work}
Although \datasetname{} creates a new benchmark for geometric protein multi-task learning, it is hard to add more tasks into \datasetname{} while maintaining the fully-labeled sample size under our definition of fully-labeled data. Relaxing the restriction of the test set can alleviate this issue. Meanwhile, the input is now a mixture of single chains and complexes, we cai randomly augment single-chain samples from their original PDB complex to form 'complexes' and label them based on their UniProt IDs. Besides, we only consider invariant tasks in this work, we can also extend our model to more tasks in future work (e.g. equivariant tasks).
\clearpage


\end{document}